# Integrating Fuzzy and Ant Colony System for Fuzzy Vehicle Routing Problem with Time Windows

Sandhya and V.Katiyar

Department of Computer Science and Engineering, Maharishi Markandeswar University, Ambala, Haryana,India.

## Abstract

*In this paper fuzzy VRPTW with an uncertain travel time is considered. Credibility theory is used to model the problem and specifies a preference index at which it is desired that the travel times to reach the customers fall into their time windows. We propose the integration of fuzzy and ant colony system based evolutionary algorithm to solve the problem while preserving the constraints. Computational results for certain benchmark problems having short and long time horizons are presented to show the effectiveness of the algorithm. Comparison between different preferences indexes have been obtained to help the user in making suitable decisions*

## Keywords:

## 1.Introduction

Transportation is one important component of logistics. Efficient utilization of vehicles directly affects the logistic cost as 40% to 50% money is spent on transportation. Moreover proper utilization of vehicles is also important from environmental view point. The use of automated route planning and scheduling can lead to huge savings in transportation cost ranging from 5% to 20% [25]. So efficient routing of fleets is a crucial issue for companies and this can be formulated as vehicle routing problem.Vehicle Routing Problem was first introduced by Danting and Rameser [26] in 1959. Till now many variants of the problem have been proposed [2]. Vehicle Routing Problem with Time Windows is one flavor of VRP. In VRPTW a least cost route from central depot to all customers is to be designed in such a way that all customers are visited by homogenous vehicles once and only once while preserving the capacity and time window constraints. In this problem it is assumed that the time to travel from one customer to another is equivalent to distance between the customers. Moreover in these types of problems, all parameters are assumed to deterministic.  In real life scenarios these assumptions donot hold because of varying road conditions, link failures, rush hours, congestion etc. which lead to uncertainties in data. Most of the algorithms developed for deterministic problems do not work in these situations. In this paper VRPTW with fuzzy travel time is considered. This uncertainty is handled using fuzzy logic. The fuzzy travel time is represented by triangular fuzzy number.

Credibility theory is used to model the problem. Improved Ant Colony System (IACS) [4] is then used to find the most efficient route for the problem. The main goal of this paper is to develop an algorithm that can provide the user with a route having minimum distance with uncertain travel time at a desired preference index. The rest of the paper is organized as follows.Literature review is presented in section 2. Section 3 discusses the essential basics of fuzzy theory. A model based on credibility theory is then proposed in section 4. Section 5 presents improved ACS that is used to solve the model. Results based on experimentations are discussed in section 6. Finally section 7 presents the conclusions and scope for future work.

## 2.Literature Review

VRP is concerned with finding the minimum set of routes, starting and ending at the central depot, for homogenous vehicles to serve number of customers with demands for a goods such that capacity constraint is preserved.Latest taxonomy on VRP can be found in [2]. The VRPTW considers the time window for each customer in which service has to be provided. A vehicle can arrive before the starting time of the window but it cannot arrive after the closing time of window. A taxonomy can be found on [5,6].VRPTW is NP-hard problem. Many exact and metaheuristics algorithms have been proposed for solving VRPTW [7,8,9,10]. A categorized bibliography of metaheuristics and their extensions can be found in [1].But these became infeasible for dynamic problems.In [11] a mixed integer linear programming approach and a nearest neighbor, branch and cut algorithm is used to solve the problem for time dependent VRPTW.However this model does not follow FIFO property.Ichoua et al [12] proposed a tabu search approach to solve TDVRPTW, where customers are characterized by soft time windows. The model presented satisfies the FIFO property.In another approach Donati et al. [13] use multi ant colony and local search improvement approach to update the slack or the feasible time delays. The travel times are analyzed by discretizing the time space thus satisfying the FIFO property. However these models fail when uncertainties arises in various parameters.A detailed summary for various uncertain parameters for VRPTW can be found in [19].Because of lack of data or due to extreme complexity of the problem it requires subjective judgment. Fuzzy set theory provides meaningful methodologies to handle uncertainty, vagueness and ambiguity.In [14] triangular fuzzy numbers are used to represent fuzzy travel time and a route construction method is proposed to solve the problem.A fuzzy optimization model using imperialist competitive algorithm is presented for fuzzy vehicle routing problem with time window in [15].Cao Erbao et al [16] use fuzzy credibility theory to model the vehicle routing problem with fuzzy demand. It uses integration of stochastic simulation and differential evolution algorithms to solve the same model. However it requires a lot of parameters to be taken care of.Yongshuang Zheng and Baodinf Liu [17] present an integration of fuzzy simulation and genetic algorithms to design a hybrid intelligent algorithm for solving fuzzy vehicle routing model. In [18] two new types of credibility programming models including fuzzy chance constraint programming and fuzzy chance-constraint goal programming are presented to model fuzzy VRP with fuzzy travel time. In [3] fuzzy concepts and genetic algorithm are used for the solution multi-objective VRP.  Finally J.Brito et al [20] proposes a GRASP meta heuristic to solve the VRPTW in which travel time is uncertain. A chance constraint model is build using credibility approach to solve the problem. However the proposed algorithm appears to be inefficient because in GRASP metaheuristic each restart is independent of the previous one.On the other hand the stochastic element of ACO allows to build variety of different solutions.In this paper VRPTW with fuzzy travel time is considered. Credibility theory is used to build a chance constraint programming model for the problem. An improved ant colony [4] metaheuristic is then used to obtain the optimal routes .Moreover using this approach a decision maker can also evaluate different planning scenarios to choose the best alternative with desired confidence level.

# 3.Fuzzy Credibility Theory

In this section, some basic concepts and results about fuzzy measure theory are summarized. The term fuzzy logic was introduced by Lofti A. Zadeh [21]. In contrast to conventional logic, fuzzy logic allows the intermediate values to be defined between conventional evaluations like true or false. Fuzzy numbers are the numbers that possess fuzzy properties. In this paper we have represented the fuzzy travel time between two customers as the triangular fuzzy number. A triangular fuzzy number is represented by triplet *TFN=(a, b, c)* where *b* is the mean value (i.e. mode) and *a* and *c* the left and right extremes of its spread. Its membership function is

$$TFN(x) = \begin{cases} 0 & x < a \\ \frac{x-a}{b-a} & a \le x \le b \\ \frac{c-x}{c-b} & b \le x \le c \\ 0 & x > c \end{cases} \qquad (1)$$

Zadeh [23] proposed the concept of possibility measure for fuzzy variables as a counterpart of probability theory in crisp sets.Concept of the fuzzy set, fuzzy variable, possibility measure, necessity theory are available in [22].Liu [24] proposed the credibility as the average of possibility and necessity. Possibilityof an event is measured by most favorable cases only in contrast to probability of an event where all favorable cases are measured. Let $\theta$ be nonempty set and P ($\theta$) be the power set. Each element in P is called an event. Also $\phi$ denotes an empty set. In order to present an axiomatic definition of possibility, it is necessary to assign a number Pos{A} to each event A, which indicates the possibility that A will occur [16].

**Axiom 3.1**: Pos($\theta$) =1 (Normality Axiom)

**Axiom 3.2**:Pos($\phi$)=0 (Non negativity Axiom)

**Axiom 3.3**: For each $A_i \wedge p(\theta)$

Pos$\{\bigcup_{i=1}^{\infty} A_i\}$=V$_{i=1}^{\infty} Pos(A_i)$      (Maximality Axiom)

Figure 2.1 shows the possibility of fuzzy event {X< = x$_0$}

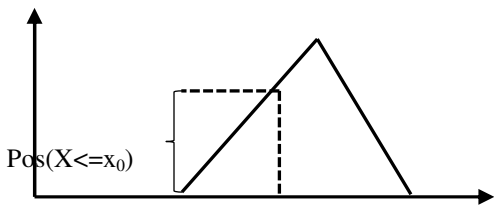

Figure 2.1 Possibility of fuzzy event

The explicit expression for possibility Pos(X<=x0) is given by:

$$\text{Pos}(X \le x_0) = \begin{cases} 1 & b < x_0 \\ \frac{x_0 - a}{b - a} & a \le x_0 \le b \\ 0 & x_0 < a \end{cases} \qquad (2)$$

Necessity for an event A is defined as the impossibility of complement of that event $A^c$ i.e Nec{A}=1-Pos {$A^c$}. Fig 2.2 shows the necessity of fuzzy event {X<=x0}

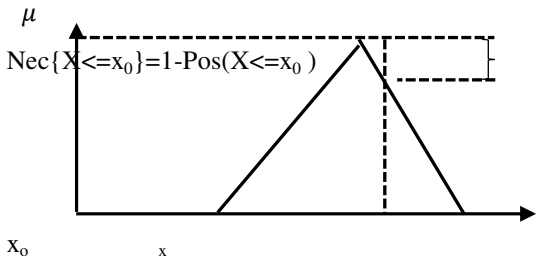

Figure 2.2 Necessity of fuzzy event

The explicit expression for possibility Nec(X<=x0) is given by:

$$\text{Nec}(X \le x_0) = \begin{cases} 1 & c < x_0 \\ \frac{x_0 - b}{b - c} & b \le x_0 \le c \\ 0 & x_0 < b \end{cases} \qquad (3)$$

The credibility for event A is Cr{A}=1/2(Pos{A}+Nec{A}).}). The credibility measure signifies the credibility how the solution satisfies the constraints. The explicit expression for credibility in case of triangular fuzzy measure is:

$$\text{Cr }(X \le x_0) = \begin{cases} 1 & c < x_0 \\ \frac{x_0 - 2b + c}{2(c - b)} & b \le x_0 \le c \\ \frac{x_0 - a}{2(b - a)} & a \le x_0 \le b \\ 0 & x_0 < a \end{cases} \qquad (4)$$

## 4. Chance Constraint Model for Fuzzy VRPTW

We define a Fuzzy VRPTW as follows:

Given a set of *n* geographically distributed customers requiring services within a specific time period from set of *k* homogenous vehicles stationed at a central depot with known demands of customers having uncertain travel time between customers but lying within known ranges with most likely values being known then the objective of fuzzy VRPTW is to find the route with minimum distance with specified confidence level that will meet all customers' time windows requirements assuming that:

1. Each vehicle starts and ends its tour at central depot indexed by 0.

2. Each vehicle has fixed identical capacity.

3. Each customer is to be visited once and only once by only one vehicle.

4. Each customer has a predefined time window in which it has to be served $[e_i , l_i]$.

5. The demand of the customer is fixed and is assumedthat it will not exceed the vehicle capacity.

6. The travel time between each customer is not fixed and it is assumed to be expressed as triangular fuzzy number $\widetilde{t_{ij}}=(t_{ij1}, t_{ij2}, t_{ij3})$.

Let $D[i]$ be the departure time of vehicle v from customer $i$, then arrival time $A[j]$at j will be the summation of departure time from previous customer i and the fuzzy travel time from node i to j.

$$A[j] = D[i] + \widetilde{t_{ij}} \qquad (5)$$

The time to begin service will be maximum of arrival time or the opening time.

$$TS[j] = max\ (A[j],\ e_j) \qquad (6)$$

Because of fuzzy travel time, arrival time and time to begin service at next customer will also be fuzzy. However the service time $S[i]$ and opening of the time windows $e[i]$ are crisp numbers that are special case of triangular fuzzy numbers where the three defining numbers are equal. We obtain the credibility that the time to begin service at next customer does not exceed its close time to be

$$Cr(TS[j] \trianglelefteq l_j) = \begin{cases} 0 & if\ l_j < TS[j]_1 \\ \frac{l_j - TS[j]_1}{2*(TS[j]_2 - TS[j]_1)} & if\ l_j \geq TS[j]_1 and\ l_j < TS[j]_2 \\ \frac{l_j - 2*TS[j]_2 + TS[j]_3}{2*(TS[j]_3 - TS[j]_2)} & if\ l_j \geq TS[j]_2 and\ l_j < TS[j]_3 \\ 1 & if\ l_j \geq TS[j]_3 \end{cases} \qquad (7)$$

As we know that if the travel time to next customer is smaller than its closing time then the chances of serving that customer by that vehicle will grow. That is greater the difference between the closing time and the maximum travel time greater is the chance to serve that customer and if this difference is small that customer may not be served. Therefore, at some preference index $Cr^*$, the solution $r$ verifies the fuzzy constraint of service times within the corresponding time window if: $Cr(TS[r] \trianglelefteq l_j) \geq Cr^*$.The goal is to determine the value of $Cr^*$ which will result in a route having minimum distance. For this stochastic simulation is done. Thus the objective of corresponding chance constraint model of fuzzy VRPTW using credibility theory is as follows:

$$F = min \sum_{k=1}^{m} \sum_{i=0}^{n} \sum_{j=0}^{n} x_{ij}^k c_{ij}^k \qquad (8)$$

subject to:

$$\sum_{k=1}^{n} \sum_{i,j=0}^{n} x_{ij}^{k} = 1 \qquad (9)$$

$$\sum_{j=1}^{n} x_{0j}^{k} = 1 \text{ and } \sum_{i=1}^{n} x_{i0}^{k} = 1 \qquad (10)$$

$$\sum_{j=0}^{n} x_{ij}^{k} - \sum_{i=0}^{n} x_{ji}^{k} = 0 \qquad (11)$$

$$\sum_{i=1}^{n} d_i \leq Q_k \qquad (12)$$

$$\sum_{i=0}^{n} \sum_{j=0}^{n} x_{ij}^{k} (\widetilde{t_{ij}} + s_{ij} + wt_i) \leq l_i \qquad (13)$$

$$\sum_{r=1}^{n} \sum_{j=0}^{n} Cr(TS[r] \leq l_j) > Cr^* \qquad (14)$$

Following indices and notations are considered:

- $i = 0,1,2,\ldots,n$ are the customer indexes with 0 denoting base station.

- $k = 1,2,\ldots,m$ are the vehicles.

- $d_i$ is the demand of the customer.

- $Q_k$ is the capacity of the vehicle.

- $c_{ij}$ cost of moving from node i to j, expressed in terms of distance from customer i to j.

- $s_{ij}$ is the service time at customer i.

- $[e_i, l_i]$ is the time window of customer i.

- $t_{ij}$ is the fuzzy travel time between i and j.

- $wt_i$ is the waiting time at customer i.

- $TS[i]$ is the fuzzy time to begin service.

A binary variable $x_{ij}^{k}$ is introduced such that $x_{ij}^{k} = 1$ if vehicle k travels directly from customer *i* to customer *j* and otherwise 0. The objective function (8) seeks to minimize the total traveled distance whereas constraint (9), specifies that every customer is visited by one vehicle and splitting of deliveries are forbidden.Eq (10), states that each tour starts and ends at depot indexed 0. Capacity constraint is preserved in Eq (12) and it ensure that demand of each customer does not exceeds vehicle capacity.Eq. (13), preserves the constraints and ensuring that sum of fuzzy travelling time to the customer, service time and waiting time are less than the closing time of customer's window. Eq (14), preserves time to begin service for route *r* is within a specified preference index.

## 5.Proposed Solution Approach

In this section, first, stochastic simulation will be used to calculate the total distance. Then ant colony heuristic is used to obtain the least cost route plan with the best value of dispatcher preference index $Cr^*$.

## a.Stochastic Simulation for additional distance

In this paper, as travel time between each customer is uncertain and they are represented as triangular fuzzy numbers, so algorithms for deterministic problems cannot be applied for the fuzzy VRPTW. Moreover real values of the travel time will be known after reaching the customer. However uncertain travel time can be considered deterministic by stochastic simulation. We summarize the algorithm as follows:

Step 1: Simulate actual travel time for each customer by following process:

Step 1.1: Generate a random number x within the left and right boundaries of a triangular fuzzy travel time between each customer and calculate its membership u(x).

Step 1.2: Generate a random number r ∈ (0,1).

Step 1.3: Compare r and u(x): if r<u(x), use x as the actual travel time, otherwise generate x and r again and compare them until they satisfy the condition r<u(x).

Step 2: Calculate total distance by moving along the planned route.

Step 3: Repeat step 1 and 2 N times and calculate the average total distance.

## b.Route Construction using Ant Colony Optimization

To obtain the best solution, enhance ant colony optimization of [4] is applied. The algorithm works as under.

Step 1: Initialize the pheromone matrix $\tau_{ij} = {}^1\!/_{c_{ij}}$ and place $m$ ants at depot and the set of customers $N_i$ as unvisited.

Step 2: Repeat while all the customers $N_i$ are not marked as visited.

Step 2.1 Start an ant and mark depot as the current location.

Step 2.2 From the current location ($i$) choose the customer ($j$) to be visited by a pseudo-random proportional rule given by:

$$j = \begin{cases} argmax \left( [\tau_{ij}]^\alpha \cdot [\eta_{ij}]^\beta \right), & if \ r \leq r_0 \\ J & otherwise \end{cases} \qquad (15)$$

where $r \in [0,1]$ is a uniformly generated random number, and $\eta_{ij}$ is the visibility of customer $j$ and defined by

$$\eta_{ij} = {}^1\!\Big/{}_{[ \, (C_{ij} + wt_j)^\gamma + (l_j - e_j)^\delta \, ]} \qquad (16)$$

Here $C_{ij}$ is the cost of travelling from node i to j which is distance from node i to j. In our case as travel time are uncertain and their values will be known after reaching the customers so it may be possible that vehicle arrives at acustomer but cannot serve it because of expiry of time window.

Also $wt_j = \begin{cases} e_j - a_j & if \ e_j > a_j \\ 0 & otherwise \end{cases}$ is the waiting time at location $j$ before service can be started and $l_j - a_j, \ a_j < l_j$, i.e. the difference between the latest arrival time $l_j$ and actual arrival time $a_j$ at customer $j$ is the measure of urgency of customer $j$ to be served.

*Also* $J \in \psi_k$ is the customer selected according to the probability $\mathcal{P}_r$ given by

$$\mathcal{P}_r = \frac{[\tau_{ij}]^\alpha \cdot [\eta_{ij}]^\beta}{\sum_{j \in \psi_k}[\tau_{ij}]^\alpha \cdot [\eta_{ij}]^\beta} \qquad (17)$$

$\psi_k$ is the set of customers which will be successfully visited from the current location by the same vehicle without violating the following time and capacity constraints.

(i) **Time constraints:** $a_j \le l_j$ i.e. arrival at $j^{th}$ customer must be up to the closing time $l_j$ of that customer with

$$a_j = max(a_{j-1}, e_{j-1}) + S_{j-1} + t_{j-1,j} \text{ and}$$

$$Cr(TS[r] \trianglelefteq l_j). \ge Cr^* \qquad (18)$$

(ii) **Capacity constraints:** $\left( x_{ij}^k d_{ij}^k \le Q_k \right) \qquad (19)$

Step 2.3 Mark ($j$) as the current location and update the set of unvisited customers. $\mathcal{N}_i \to \mathcal{N}_i - j$. Also update capacity and current time of the ant $k$.

Step 2.4 From the current location again find $\psi_k$. If $\psi_k = \mathcal{N}ull$ then go to 2.1 else repeat the 2.2.

Step 3: Out of $m$ paths find the best path to be followed with minimum total travelled distance and further try for improving that route by applying local search.

Step 4: Update the pheromone matrix by the global pheromone updation rule:

$$\tau_{ij} = (1 - \rho)\tau_{ij} + \sum_{k=1}^{K} \Delta \tau_{ij} \qquad (20)$$

where $\rho \in (0, 1)$ is a constant that controls the speed of evaporation of pheromone. $K$ is number of routes in the current best solution. The deposited pheromone $\Delta \tau_{ij}$ on the links is given by $\Delta \tau_{ij} = Q/L$. Here $Q$ is the constant, $L$ is the tour length of current solution.

Step 5: Repeat the step 2, 3 for desired number of iteration and provide the best obtained solution.

## 6. Computational Experience

The proposed algorithm has been encoded in MATLAB 8.0. This study uses the dataset of J.Brito et al [20] which was generated from the example of Zheng and Liu [17]. Two types of experimental conditions based on the time windows are generated. We assume that there are 18 customers with short time horizons and with long time horizons. Customer labeled 0 is assumed as depot.  In each experiment demand for each customer, distance and fuzzy travel time between each customer is same as that of J.Brito et al. Start time and the closing time for each customer in long time horizon is assumed to be same but for dataset with short time period the total opening duration is assumed to be 100 for each customer. The relative parameters and their values that were used during implementation are listed in table6.1 .

Table 6.1 Simulation Parameters

| Sr. No. | Simulation Parameter | Parameter Value |
|---------|----------------------|-----------------|
| 1 | Number of customers | 18 |
| 2 | Number of iterations | 1000 |
| 3 | Capacity of Vehicles(Q) | 1000 |
| 4 | Initial Pheromone value for all arcs | 1 |
| 5 | A | 2.5 |
| 6 | B | .7 |
| 7 | Γ | .5 |
| 8 | P | 0.2 |
| 9 | $q_0$ | 0.5 |
| 10 | No. of Ants | 11 |
| 11 | Credibility | .8 |
| 12 | Depot Close time | 5000 |
| 13 | Service Time | 15 |

Comparison of results obtained by our algorithm,J.Brito[20] and Zheng et al [17] on various parameters are presented in table 6.2

Table 6.2 Comparison Table

| Factors | Zheng et al[17] | J.Brito et al[20] | Our Approach |
|---|---|---|---|
| Meta Heuristics | Genetic | GRASP | ANT COLONY SYSTEM |
| Total Iterations | 10,000 | - | 1000 |
| Time Consumed | 10 hours | 1 minute | 3 minutes |
| Vehicle used | 4 | 3 | 3 |
| Total Distance | 457.5 | 365.5 | 373 |
| Vehicle Routes | R1:0-16-17-18-5-0<br><br>R2:0-10-12-13-14-15-8-0<br><br>R3:0-1-2-3-0<br><br>R4:0-9-6-7-11-4-0 | R1:0-17-18-16-15-14-12-13-0<br><br>R2:0-2-1-3-4-6-8-0<br><br>R3:0-10-9-11-7-5-0 | R1:0-1- 2-3-18-17-13– 8-0<br><br>R2:0-4-5-6-10-11-7-9-0<br><br>R3:0-16-15-14-12-0 |
| Loads of Vehicles | 590,815,440,640 | 795,930,760 | 955,840,550 |
| Robust | Less | More | More |

One can observe that our algorithm produces effective results than Zheng et al [17]and comparable results with GRASP. Moreover proposed algorithm is more robust than [17] in terms of utilization of vehicles. To evaluate the importance of dispatcher preference Cr varied with the interval of 0 to 1 with a step of 0.1. The average computational results of 10 times are given in fig 6.1

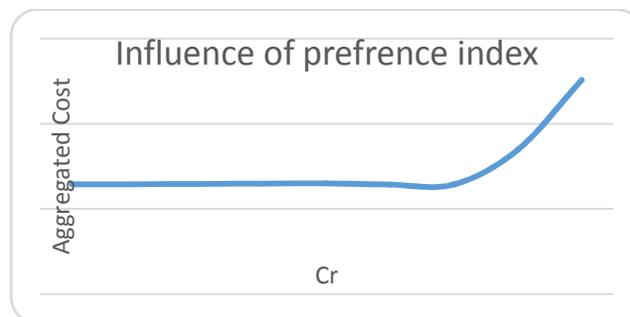

Fig 6.1 Influence of preference index

From above figure on can conclude that if decision maker is risk lover it will choose lower values of Cr whereas if he is risk adverse he will go for higher values with plan having higher cost. Figure 6.2 shows the effect of fuzzy travel time on time windows for long duration problems with a confidence level of 0.8.

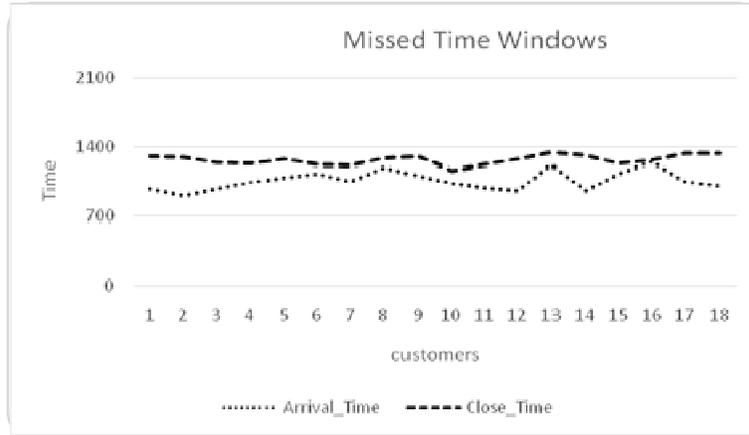

Fig 6.2 Missed time windows for long horizon problems with Cr=0.8

One can note that if we are travelling with the same fuzzy speed as given in [19] all customers are served within their time windows. But for short horizon problems if Cr<=0.4 some customers are suffered as shown in fig 6.3. For example for customer 4 and 5 the closing time window is around 900 but the actual arrival time of the vehicle is around 1000.

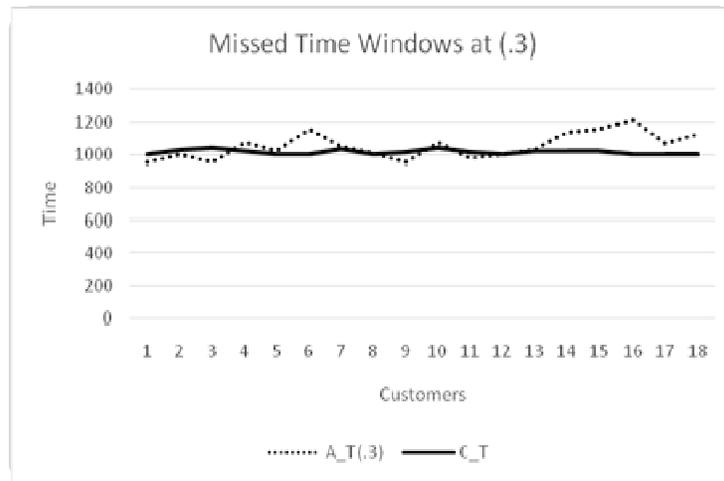

Fig 6.3 Missed time windows for short problems with Cr=0.3

If we set the Cr>=0.5 every customer can be served within its time boundaries as shown in fig6.4

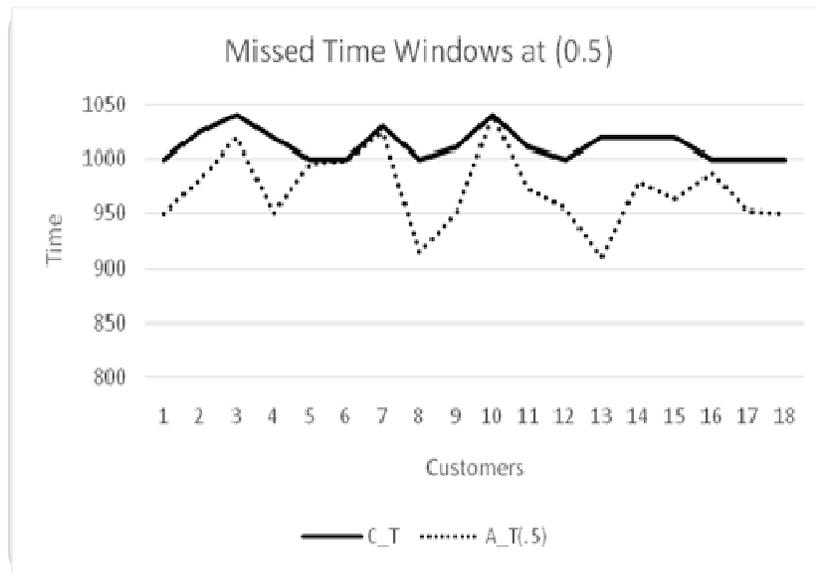

Fig 6.4 Missed time windows for short problems with Cr=0.5

# 7.Conclusions

Deterministic assumptions of the VRPTW make it unsuitable for real world environment. In this paper, VRPTW with fuzzy travel time is considered to capture the real life scenario. Our goal is to construct efficient and reliable routes for the problem. We propose a chance constrained model for the problem using fuzzy credibility theory where fuzzy travel time is represented as triangular fuzzy number. Additionally, stochastic simulation is done to get the fuzzy travel time and then ant colony optimization algorithm is used to get the optimal solution for the problem in reasonable time. We apply our solution approach to problems having short and long duration of time windows. It was concluded that the proposed approach performs well in each structure of the problem and provides improved results on [17] and comparable results with [20] approach.Further comparison between different confidence levels is done to show the influence on total distance and it was concluded that higher values of preference index leads to higher cost. This comparison helps a decision maker in choosing different values for confidence level to get different results based on cheapness and robustness.